\documentclass[conference, letterpaper]{IEEEtran}
\ifCLASSINFOpdf
\else
\fi
\usepackage{graphicx} 
\graphicspath{{../ParamOpt/}}
\usepackage{multirow}
\usepackage{url}
\usepackage{listings}

\usepackage[cmex10]{amsmath}
\usepackage{color}
\usepackage{fancyhdr}
\usepackage[caption=false,font=footnotesize]{subfig}

\renewcommand{\thispagestyle}[2]{} 

\fancypagestyle{plain}{
        \fancyhead{}
        \fancyhead[C]{first page center header}
        \fancyfoot{}
        \fancyfoot[C]{first page center footer}
}
\begin{document}

%
\title{
Characterizing the hyper-parameter space of LSTM language models for mixed context applications
}

\author{\IEEEauthorblockN{Victor Akinwande}
\IEEEauthorblockA{\\
\\
}
\and
\IEEEauthorblockN{Sekou L. Remy}
\IEEEauthorblockA{\\
\\
}}


%


\maketitle

\begin{abstract}

Applying state of the art deeplearning models to novel real world datasets gives a practical evaluation of the generalizability of these models. 
Of importance in this process is how sensitive the hyper parameters of such models are to novel datasets as this would affect the reproducibility of a model.
We present work to characterize the hyper parameter space of an LSTM for language modeling on a code-mixed corpus. We observe that the evaluated model shows minimal
sensitivity to our novel dataset bar a few hyper parameters. 

\end{abstract}


\begin{IEEEkeywords}
Language Models; Hyper parameter search; Code-mixed language
\end{IEEEkeywords}

%
\IEEEpeerreviewmaketitle

\section{Introduction}
Hyper parameter tuning is an integral part of building deep learning models.
State of the art models are often benchmarked on a small set of datasets such as Penn Treebank \cite{mikolov2010recurrent}, WikiText, GigaWord, MNIST, CIFAR10 to name a few of the limited set.
The hyper parameters values on these datasets are however not directly applicable to other use case specific datasets.

Advances in deep learning research including its applications to Natural Language Processing (NLP) is correlated to the introduction of new increasing strategies for regularization
and optimization of neural networks. These strategies, more often than not introduce new hyper parameters, thus, compounding the challenge of hyper parameter tuning; even more so if
hyper parameter values are overly sensitive to the dataset. The effect of this would be that reproducing state of the art neural models on a unique dataset would require
significant hyper parameter search thus limiting the reach of these models to parties with significant computing resources.

We present work done to understand the effect of the set of parameters selected on the perplexity (the exponential of the average negative log-likelihood of prediction of the next word in a sequence\cite{bengio2003neural}) of a Neural Language Model (NLM).
We apply hyper parameter search methods given baseline hyper parameter values for benchmark datasets to modeling code-mixed text.
Code-mixed text is text which draws from elements of two or more grammatical systems \cite{wiki:codemix}).
Code-mixed text is common in countries in which multiple languages co-exist.
In this work we assess the performance of the AWD-LSTM model\cite{merity2017regularizing} for language modeling to better understand how relevant the published hyper parameters
may be for a code-mixed corpus and to isolate which hyper parameters could be further tuned to improve performance.

Our results show that as a whole, the set of hyperparameters considered the best\cite{merity2017regularizing} are reasonably good, however ther are better sets hyperparamers for the code-mixed corpora.
Moreover, even with the best set of hyper parameters, the perplexity observed for our data are significantly higher (i.e. performance is worse at the task of language modeling) than the performance demonstrated in the literature.
Finally, our implemented approach is one that not only enables confirmation of the goodness of the hyper parameters values, but we can also develop inferences about which hyper parameter values would yield better results.


\section{Background}
\begin{table*}[ht]
\renewcommand{\arraystretch}{1.3}
\caption{ Default hyper parameter values.
}
\label{defaultvalues}
\centering
\begin{tabular}{|c|c|c|c|c|c|c|c|c|c|c|}
\hline
emsize & nhid & nlayers & dropout & dropoute & dropouth & dropouti & wdrop & bptt & clip & lr \\
\hline
\hline
300 & 1150 & 3 & 0.4 & 0.1 & 0.3 & 0.65 & 0.5 & 70 & 0.25 & 30 \\
\hline
\end{tabular}
\end{table*}

\begin{table}[ht]
\renewcommand{\arraystretch}{1.3}
\caption{ Hyper parameter space.
}
\label{searchspace}
\centering
\begin{tabular}{|c|c|}
\hline
Hyper parameter & Potential value\\
\hline
\hline
emsize & [300, 350, 400, 450] \\
\hline
nhid & [950, 1050, 1150, 1250] \\
\hline
nlayers & [2, 3, 4, 5] \\
\hline
dropout & [0.3, 0.4, 0.5, 0.6]\\
\hline
dropoute & [0.3, 0.4, 0.5, 0.6]\\
\hline
dropouth & [0.3, 0.4, 0.5, 0.6]\\
\hline
dropouti & [0.3, 0.4, 0.5, 0.6]\\
\hline
wdrop & [0.3, 0.4, 0.5, 0.6]\\
\hline
bptt & [50, 60, 70, 80]\\
\hline
clip & [0.15, 0.25, 0.35, 0.45]\\
\hline
lr & [20, 30, 40, 45]\\
\hline
\end{tabular}
\end{table}


Deeplearning has found sucess in various applications including natural language processing tasks such as language modeling, parts of speech tagging, summarization and many others.
The learning performance of deep neural networks however depends on systematic tuning of the hyper parameters. As such finding optimal hyper parameters is an integral part of building neural models
including neural language models.

Recurrent neural networks (RNNs) being well suited to dealing with sequences of vectors, have found much success in NLP by leveraging the sequential structure of language,
A variant of RNNs known as Long Short-Term Memory Networks (LSTMs) \cite{hochreiter1997long} have particularly been widely used and stands as the state of the art technique for language modeling on
benchmark datasets such as Penn Treebank (PTB) \cite{mikolov2010recurrent} and One billion words \cite{chelba2013one} among others.
Language models (LMs) by themselves are valuable because well trained LMs improve the underlying metrics of downstream tasks such as word error rate for speech recognition,
BLEU score for translation. In addition, LMs compactly extract knowledge encoded in training data \cite{jozefowicz2016exploring}.

The current state of the art on modeling both PTB and WikiText 2 \cite{merity2016pointer} datasets as reported in \cite{merity2017regularizing} shows little sensitivity to hyper parameters;
sharing almost all hyper parameters values between both datasets. In \cite{kaiser2017one}, its is also shown that deep learning model can jointly learn a number of
large-scale tasks from multiple domains by designing a multi-modal architecture in which as many parameters as possible are shared.


Training and evaluating a neural network involves mapping the hyper parameter configuration (set of values for each hyper parameter) used in training the network to the validation error obtained at the end.
Strategies for searching and obtaining an optimal configuration that have been applied and found considerable success include grid search, random search, Bayesian optimization, Sequential Model-based Bayesian Optimization (SMBO) \cite{brochu2010tutorial},
deterministic hyperparameter optimization methods that employs radial basis functions as error surrogates proposed by \cite{ilievski2017efficient}, Gaussian Process Batch Upper Confidence Bound (GP-BUCB) \cite{desautels2014parallelizing};
an upper confidence bound-based algorithm, which models the reward function as a sample from a Gaussian process.
In \cite{brazdil2008metalearning}, the authors propose initializing Bayesian hyper parameters using {\bf meta-learning}. The idea being initializing the configuration space for a novel dataset based on configurations that are known to perform well on similar, previously evaluated, datasets.

Following a meta-learning approach, we apply a genetic algorithm and a sequential search algorithm, described in the next section, initialized using the best configuration reported in \cite{merity2017regularizing} to search the space around optimal hyper parameters for the AWD-LSTM model.
Twitter tweets collected using a geo-location filter for Nigeria and Kenya with the goal of acquiring a code-mixed text corpus serve as our evaluation datasets. 
We report the test perplexity distributions of the various evaluated configurations and draw inferences on the sensitivity of each hyper parameter to our unique dataset.

\section{Methodology}

We begin our work by establishing what the baseline and current state of the art model is for a language modeling task \cite{merity2017regularizing}.
Applying the AWD-LSTM model, based on the open sourced code and trained on code-mixed Twitter data, we sample 84 different hyper parameter configurations for each dataset,
and evaluate the resulting test perplexity distributions while varying individual hyperparameter values to understand the effect of the set of hyper parameter values selected on the model perplexity.

\subsection{Datasets}

Two sources of data are collected using the Twitter streaming API with a geolocation filter set to geo-cordinates for Kenya and Nigeria. 
The resulting data is code-mixed with the Kenya corpus (Dataset 1) containing several mixes of English and Swahili both official languages in Kenya.
The Nigeria data (Dataset 2) on the other hand, does not predominantly contain mixes of English with another language in the same sentence. 
Rather, English is simply often completety rewritten in a pidgin form. 
The training data for Kenya and Nigeria contains 13,185 words and 27,855 words respectively. All tweets are stripped of mentions and hashtags as well as converted to lower-case. 

The phenomenon of code-mixed language use is common in locales that are surrounded by others which speak different languages or locales with a large number of immigrants.
In Kenya and Nigeria as such, the use of English is influenced by the presence of one or more local languages and this is evident in the corpus.

\subsection{Model Hyper parameters}

We considered 11 hyper parameters for tuning including the size of the word embedding (emsize), the number of hidden units in each LSTM layer (nhid), the number of LSTM layers (nlayers), the initial learning rate of the optimizer (lr),
the maximum norm for gradient clipping (clip), the backpropagation through time sequence length (bptt), dropout - applied to the layers (dropout), weight dropout applied to the LSTM hidden to hidden matrix (wdrop), the input embedding
layer dropout (dropouti), dropout for the LSTM layer (dropouth), and dropout to remove words from embedding layer (dropoute).
Table \ref{defaultvalues} contains the default values of the individual hyper parameters. 

All experiments involved training for 100 epochs inline with available GPU resources. 
The training criteron was the cross-entropy loss which is the average negative log-likelihood of predicting the right next word by the LM.
It took approximately two hours wall clock time to train the model for each hyper parameter configuration.

\subsection{Sequential search}

The search process begins by setting the values of each hyper parameter (configuration) to known best values (see Table \ref{defaultvalues}).
We then iteratively search for the best value for each hyper parameter.
The order used in this search is defined in the rows of Table \ref{searchspace}.
Performance is evaluated based on the text perplexity for the modeling task.
Once the best perplexity is identified from the list of possible values for the given hyper parameter, it is fixed and the space of the next hyper parameter in the sequence is searched.
In this manner the the configuration space of the model is explored.

This approach shares similarities with the method applied in \cite{wang2016semi}, though it remains an open question what the impact of the sequence is on the quality of best configuration produced.
For the context of this work, our aim is not to find the best configuration.
Instead it is to better understand the configuration space defined by these hyper parameters to determine the impact of their values on the performance when considering a code-mixed corpora.

\subsection{Population based search}
We apply a genetic algorithm (GA) to provide a complementary approach to the sequential search for the exploration of hyper parameter configurations. 
The GA is a biologically inspired population based search technique presented by \cite{holland1992genetic}. The algorithm is a meta-heuristic inspired by 
biological natural selection. The test perplexity for a particular hyper parameter configuration is the measure of its fitness.
Given an evaluated population i.e a set of hyper parameter configurations, we derive the next generation of the population by first selecting the candidate
configurations using roulette wheel selection \cite{goldberg1991comparative}. This biases the selection of good configuration to pass their "genetic material"
to the subsequent generation. The probability of selection of the $ jth $ configuration in a generation, $ p^j, $ is defined in (\ref{eq:1}) where $ f^j $ is its
fitness which is a function of the test perplexity.
\begin{equation} \label{eq:1}
p^j = f^j/\sum_{k=1}^{B} f^k
\end{equation}
Each hyper parameter configuration in the subsequent generation is derived from two parent configurations selected via this approach. Mimicking biological crossover chromosomes \cite{holland1992genetic},
the two configurations selected are mixed, and one of the resulting configurations are selected at random. Finally, a random subset of the components of 
each derived configuration is perturbed by adding noise. This sequence of processes define how configurations from a current generation are used to derive the next generation.
\subsection{Meta-learning initialization}
Both the population based and sequential search space were manually initialized with four (4) values of each hyper-parameter in the neighbourhood of
the best values reported in \cite{merity2017regularizing} as shown in Table \ref{defaultvalues}.
It is important to note that the sampled configuration space is very small compared to the overall space which is of size $ 4^{11} $.
84 samples for each dataset constitute the set of sampled configurations which is a far cry from the size of the universal set.

\section{Results}
We use the term default value when refering to an individual hyper parameter value that makes up the configuration with the best result for the AWD-LSTM model as reported in \cite{merity2017regularizing}.
We evaluate the sensitivity of the hyper parameters by observing the test perplexity distribution comparing it with the default values.

\subsubsection{Better}

When considering each of these hyper parameters, it's possible to identify that the default value is correlated with a statistically better performance.
This is the case for dropouti, dropoute, dropouth, clip, wdrop, and lr.

\subsubsection{Not better but not worse! (though generally the best)}

\begin{figure}[t]
  \center{
  \subfloat[][Dataset 1]{
  \includegraphics[width=.48\columnwidth]{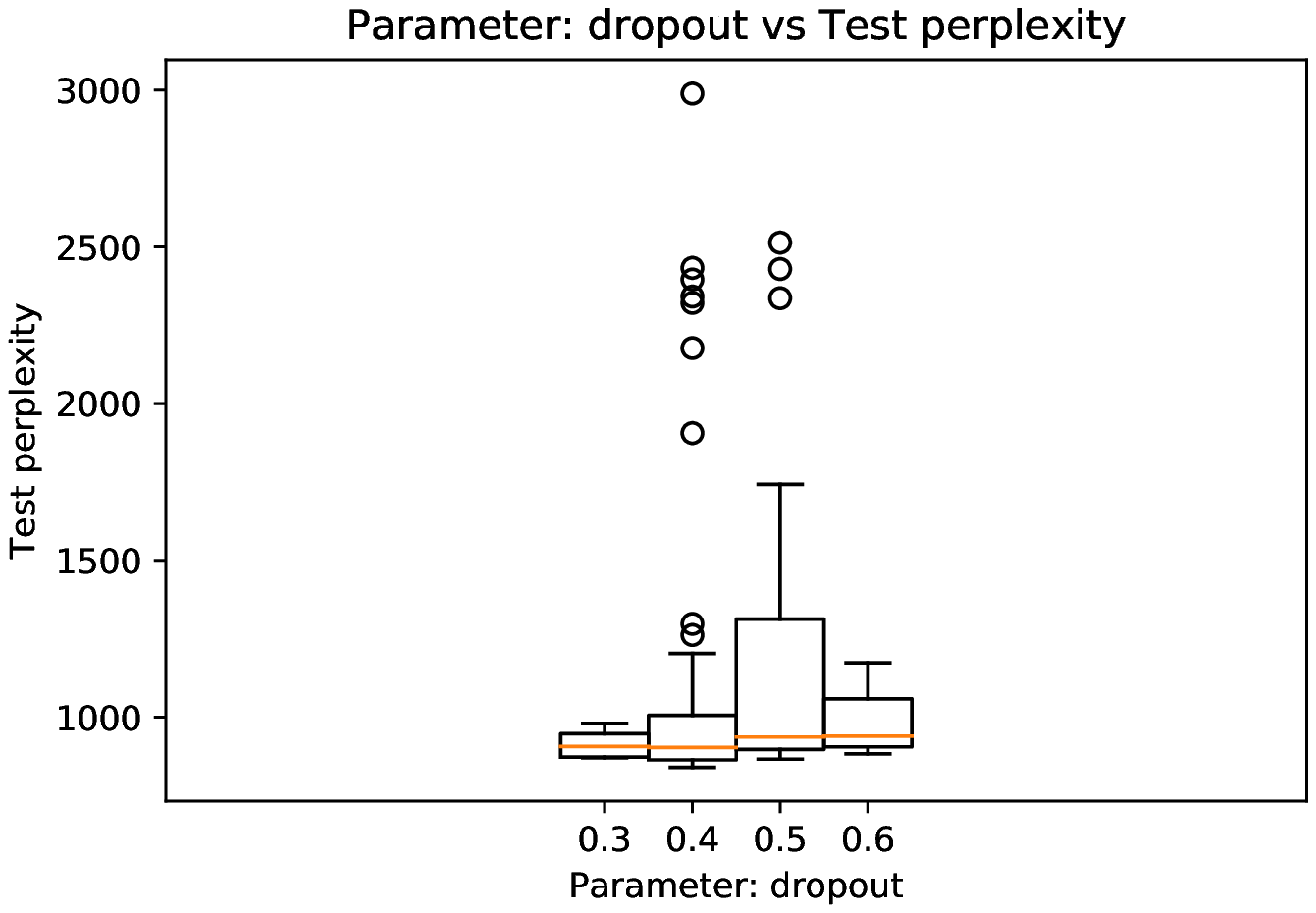}
  \label{fig:dataset1dropout}
  }
  \subfloat[][Dataset 2]{
  \includegraphics[width=.48\columnwidth]{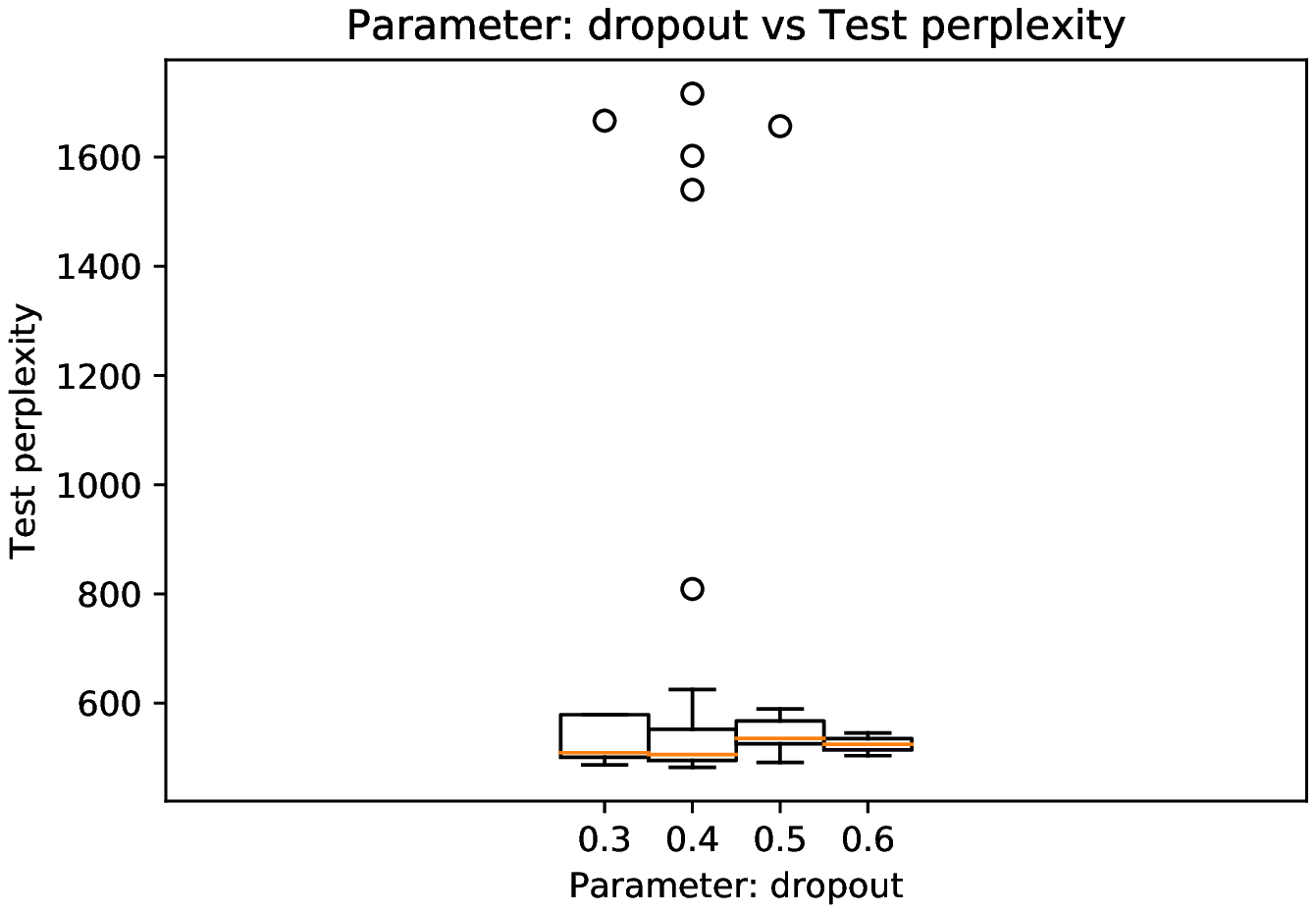}
  \label{fig:dataset2dropout}
  }
  \caption{Boxplot of the perplexity as a function of the dropout}
  \label{fig:dropout}
}
  \end{figure}

\begin{figure}[t]
    \center{
    \subfloat[][Dataset 1]{
    \includegraphics[width=.48\columnwidth]{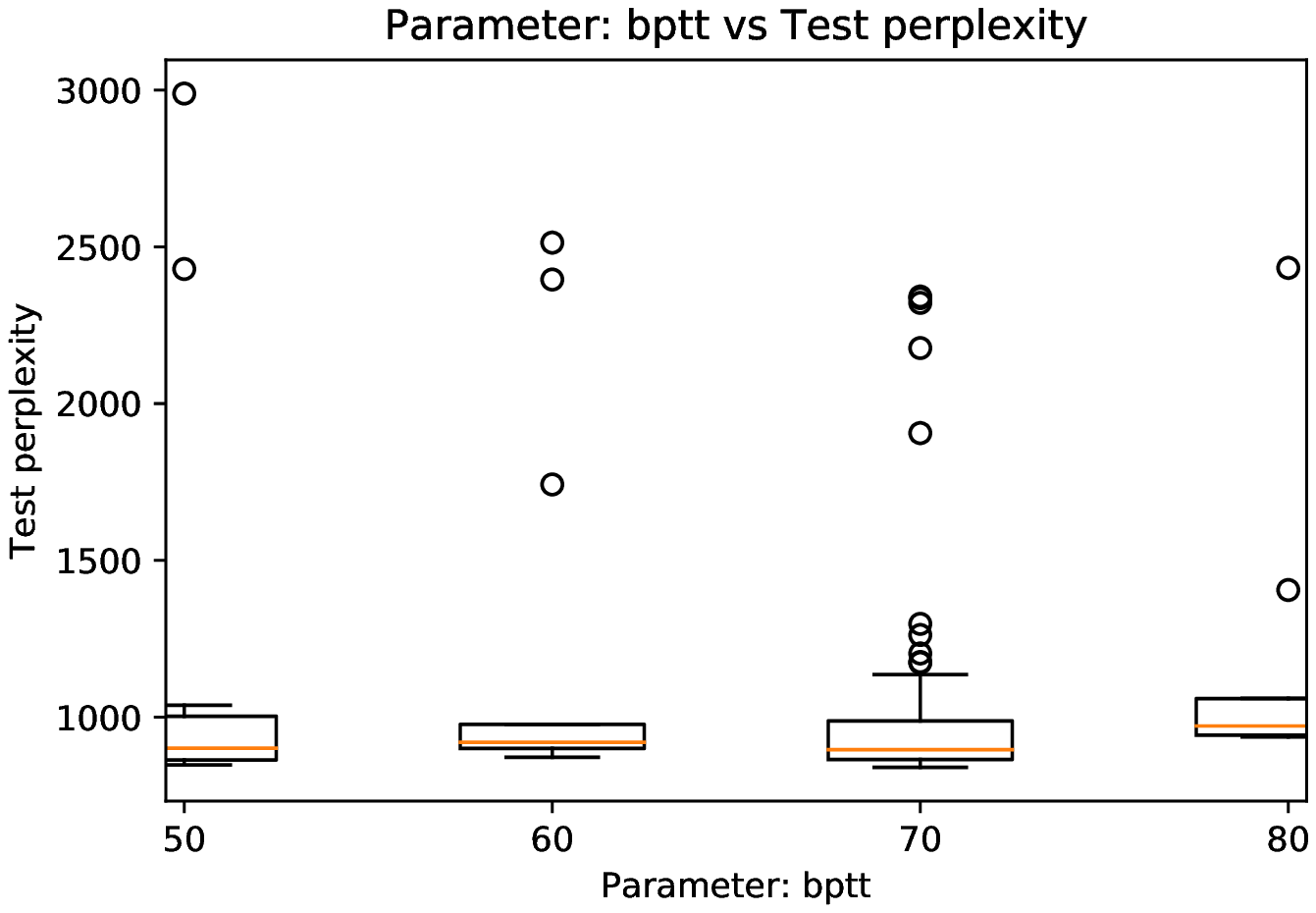}
    \label{fig:dataset1bptt}
    }
    \subfloat[][Dataset 2]{
    \includegraphics[width=.48\columnwidth]{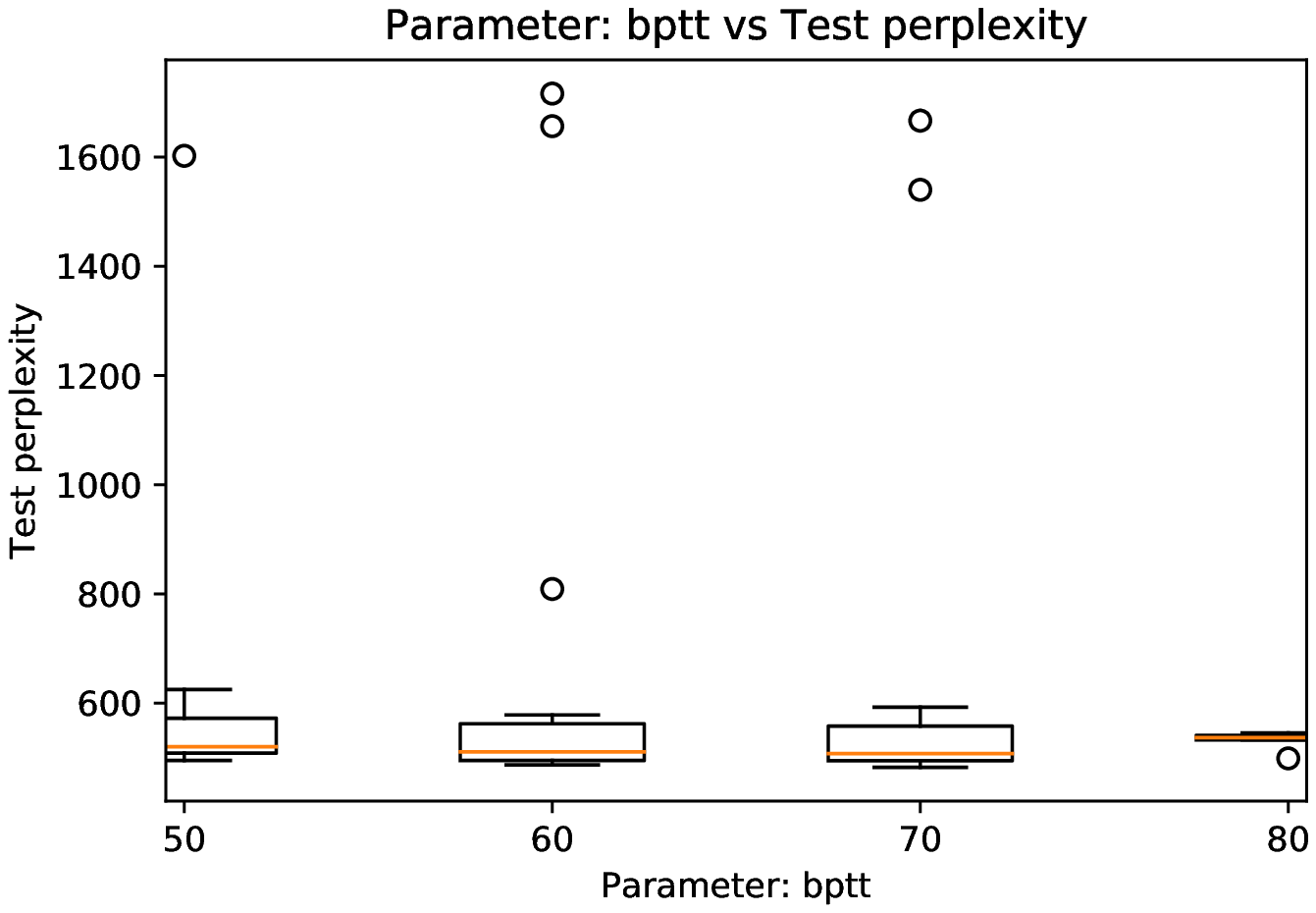}
    \label{fig:dataset2bptt}
    }
    \caption{Boxplot of the perplexity as a function of the bptt sequence length}
    \label{fig:bptt}
}
    \end{figure}

When considering each of these hyper parameters, it's possible to identify that the default value is not correlated with a statistically better performance but also not correlated with statistically worse performance. 
However the default values are closely in the neighbourhood of the best hyper parameter values for both datasets.
This is the case for dropout and bptt as shown in \ref{fig:dropout} and \ref{fig:bptt} respectively.

\subsubsection{Not better but not worse! (though generally not the best)}

\begin{figure}[t]
  \center{
  \subfloat[][Dataset 1]{
  \includegraphics[width=.48\columnwidth]{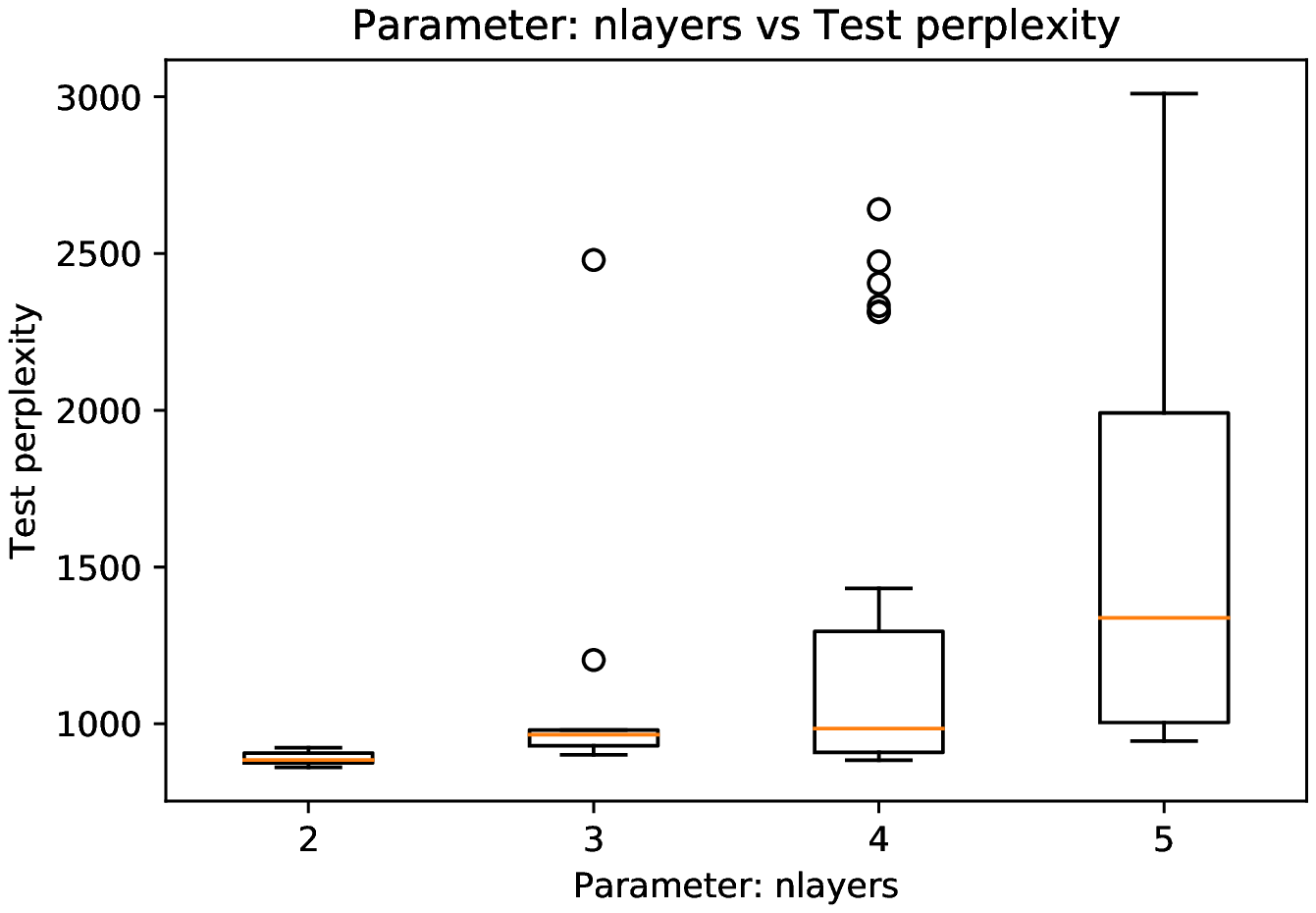}
  \label{fig:dataset1nlayers}
  }
  \subfloat[][Dataset 2]{
  \includegraphics[width=.48\columnwidth]{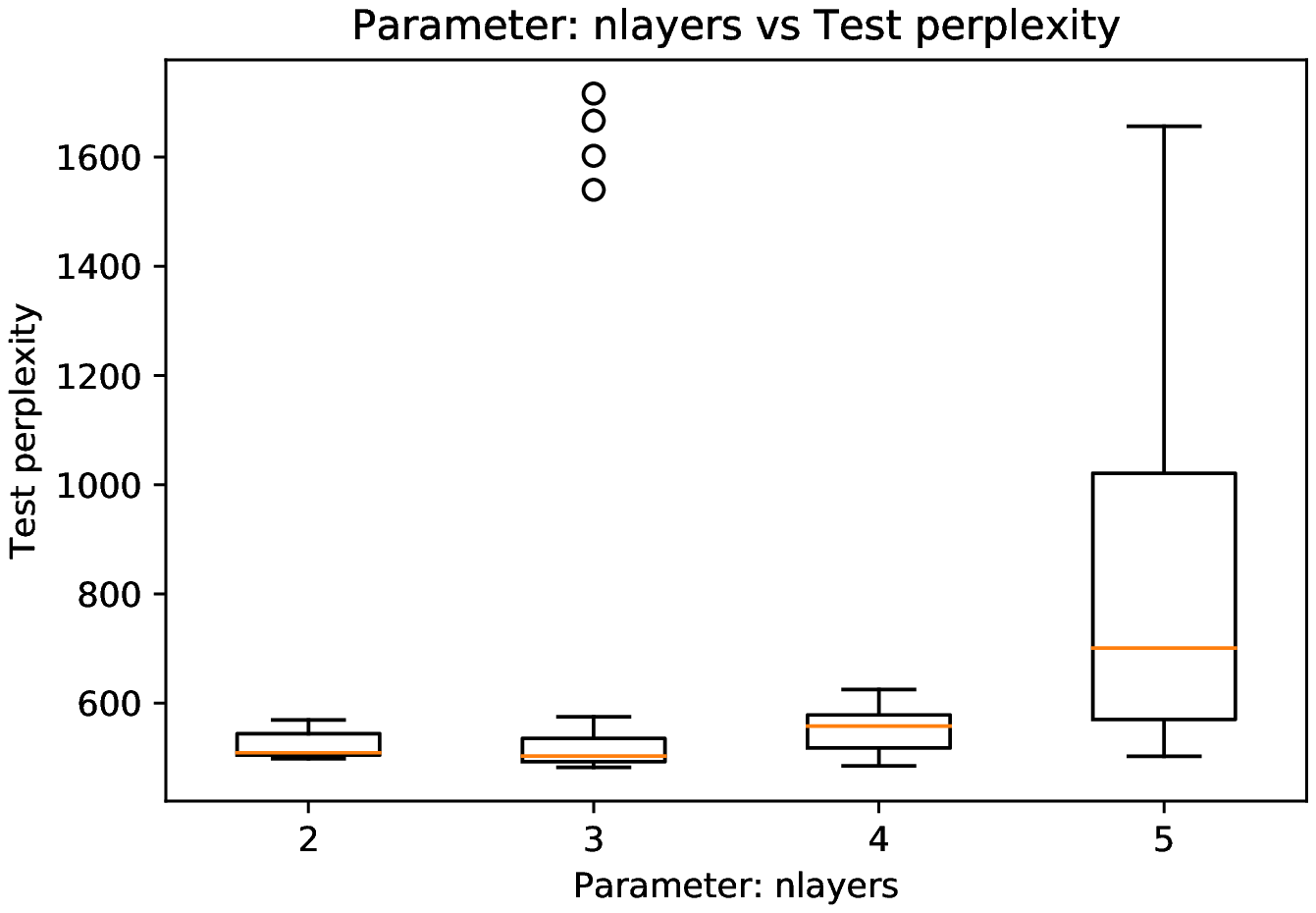}
  \label{fig:dataset2nlayers}
  }
  \caption{Boxplot of the perplexity as a function of the number of layers in the model.}
  \label{fig:nlayers}
    }
  \end{figure}

\begin{figure}[t]
  \center{
  \subfloat[][Dataset 1]{
  \includegraphics[width=.48\columnwidth]{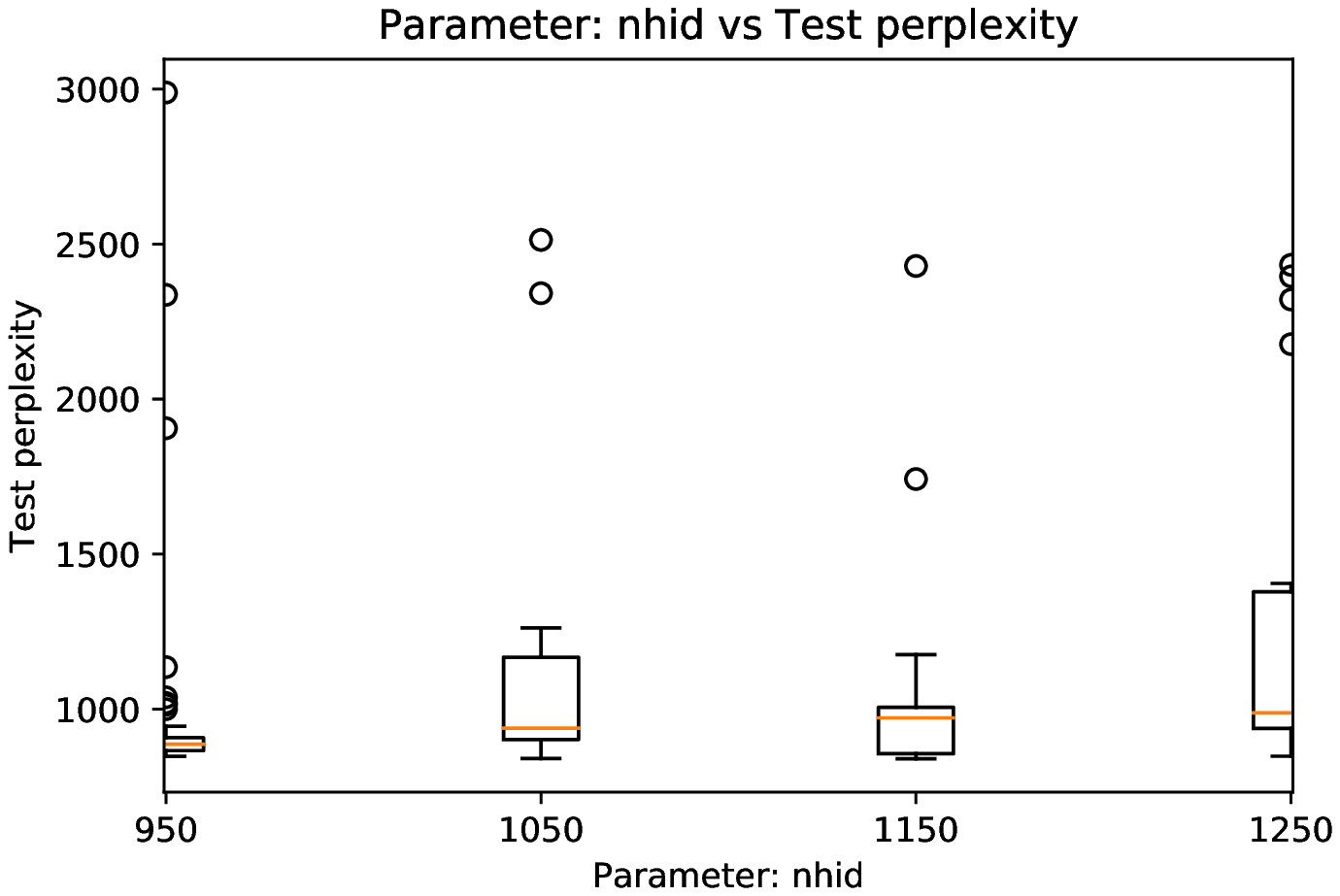}
  \label{fig:dataset1nhid}
  }
  \subfloat[][Dataset 2]{
  \includegraphics[width=.48\columnwidth]{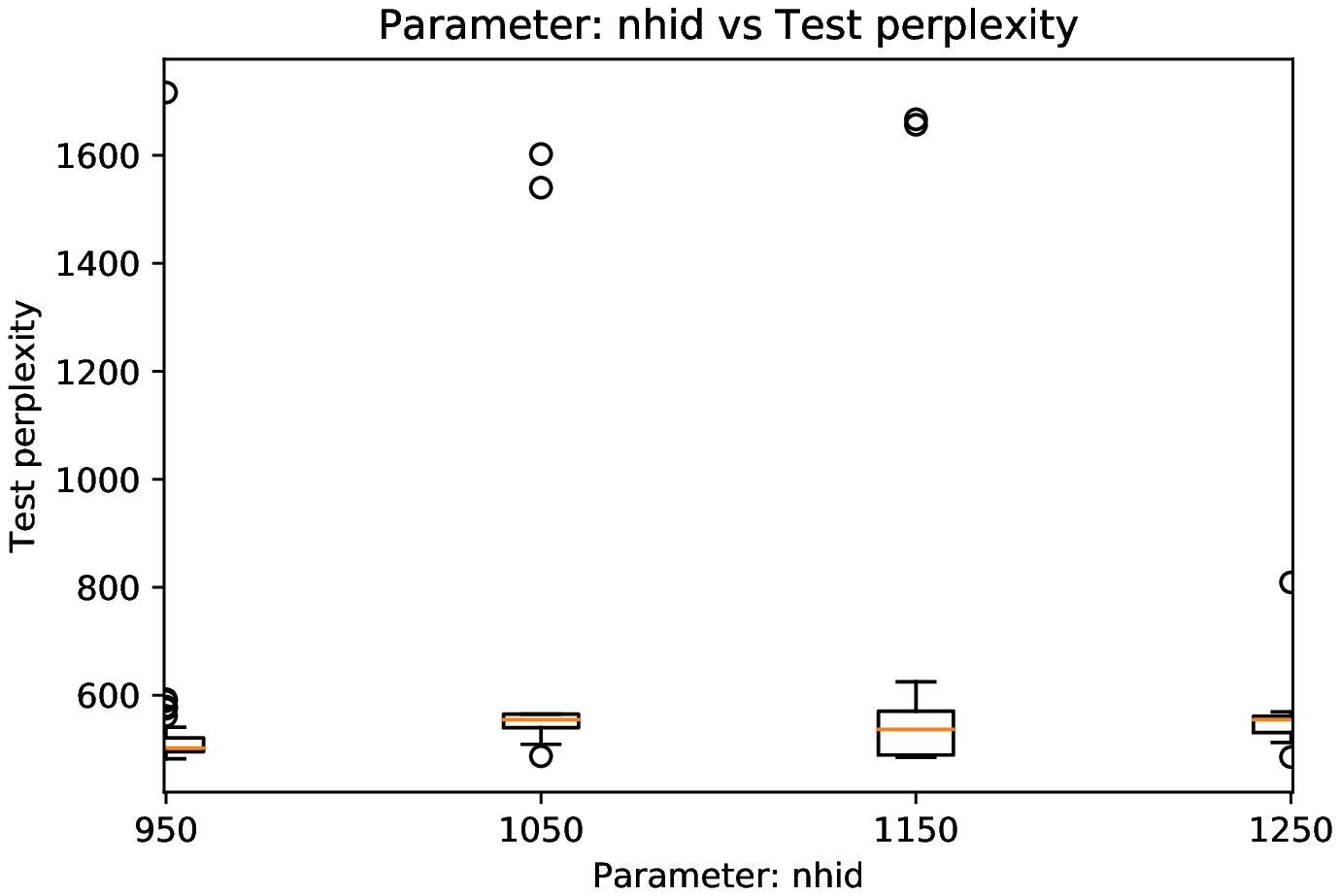}
  \label{fig:dataset2nhid}
  }
  \caption{Boxplot of the perplexity as a function of the number of hidden units in the model.}
\label{fig:nhid}}
  \end{figure}

When considering each of these hyper parameters, it's possible to identify that the default value is not correlated with a statistically better performance but also not correlated with statistically worse performance. 
However the default values are not closely in the neighbourhood of the best values. 
This is the case for the nlayers and nhid hyper parameters as shown in \ref{fig:nlayers} and \ref{fig:nhid} respectively.

The data from varying the number of LSTM layers suggests that shallow models yields lower perplexity. And this is consistent across both datasets and supported by \cite{zhang2003state}.
The number of hidden units indicates a bound on the number of nonlinear transformations in the network with an increase leading to an increase in the number of calculations between inputs and corresponding outputs.
A higher number of hidden units is expected to improve the model performance. What is observed however on both datasets is the lowest value of hidden units yielding the best result.

\subsubsection{Worse}

\begin{figure}[t]
\center{
\subfloat[][Dataset 1]{
\includegraphics[width=.48\columnwidth]{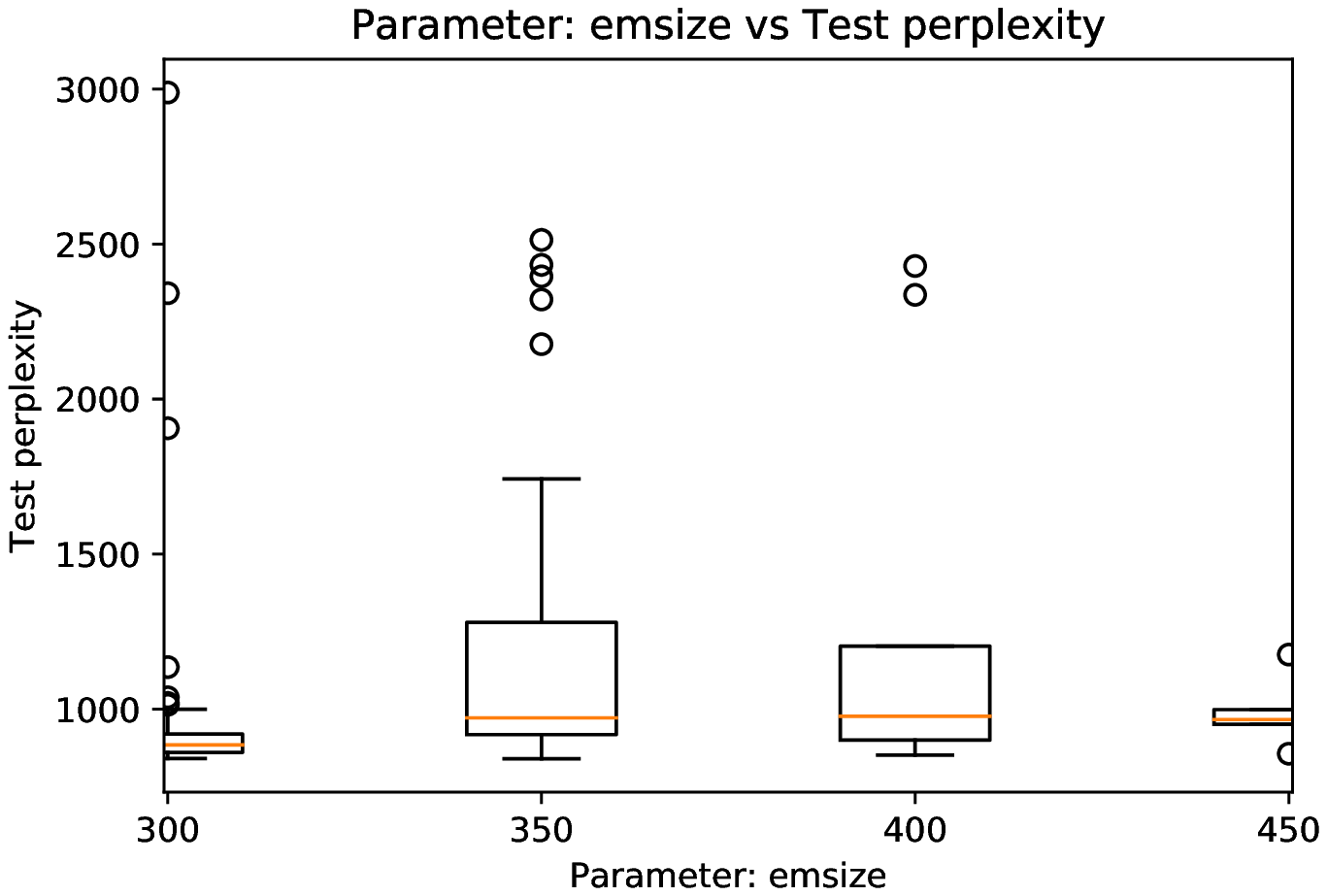}
\label{fig:dataset1emsize}
}
\subfloat[][Dataset 2]{
\includegraphics[width=.48\columnwidth]{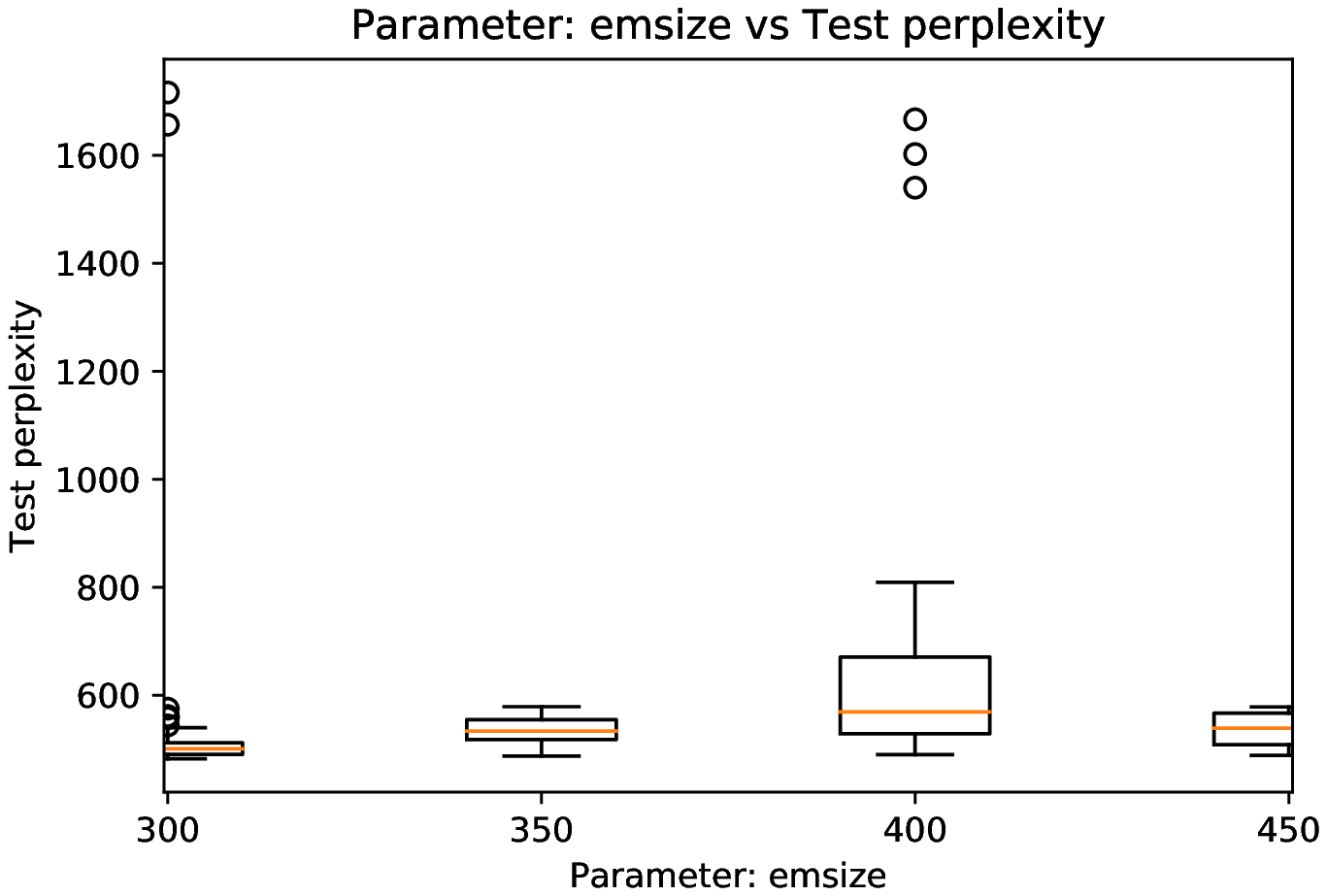}
\label{fig:dataset2emsize}
}
\caption{Boxplot of the perplexity as a function of the embedding size used for the model.}
\label{fig:emsize}
}
\end{figure}

The emsize is observed to be the only hyperparameter for which the default values is correlated with statistically significant worse performance on both Datasets 1 and 2 as shown in Figure \ref{fig:emsize}.

Encoding context, morphology, relationships between words through training the word embedding is directly tied to the vocabulary. 
As there are varying degree of similarity between words across corpora the embedding size is a hyper-parameter that is expected to be sensitive to the dataset.
Every corpus has varying level of semantic and syntatic context that needs to be encoded as features that affects the NLM. Thus, the sensitivity of the embedding size hyper-parameter is not overall surprising.

We present a comparison of the test perplexity using the default values of the AWD-LSTM model with the best test perplexity from the both sequential and population based search in Table \ref{results_comparison}.

\begin{table}[t]

\renewcommand{\arraystretch}{1.3}
\caption{Comaprison of test perplexity with default hyper parameter values defined by the state of the art \cite{merity2017regularizing} 
}
\label{results_comparison}
\centering
\begin{tabular}{|c||c||c|c|}
\hline
dataset & default values perplexity & best perplexity & changed parameters \\
\hline
Dataset 1 & 851.24 & 839.56 & emsize\\
\hline
\hline
Dataset 2 & 490.28 & 482.41 & emsize, nhid\\
\hline
\end{tabular}
\end{table}


\section{Conclusion}
In this work we set out to characterize the space of hyper parameter values for a neural language model trained to perform the task of language modeling.
The performance of such models is sensitive to the selection of hyper parameters which define their operation.
Our work applied language modeling to the domain of code-mixed text, and we found that although the published hyper parameters the performance of the state of the art architecture, the AWD-LSTM model, were largely good, they did not define the best combination of hyper parameters for the task.
Our hyper parameter searches uncovered that the AWD-LSTM model is not generally sensitive to novel datasets.
Specifically, the size of the word embedding, and the number of hidden units in each LSTM layer are observed to be the only two hyper parameters from the set of 11 evaluated hyper parameters that differ from the published work.
This work thus can serve as a solid baseline model derivation of better sets of hyper parameters for this type of data.

Of particular interest to us is the performance of the AWD-LSTM model which is the current state of the art on a code-mixed corpus.
The perplexity values are a far cry from what is generally known to be `good' perplexity of NLMs.
As such, while the AWD-LSTM model shows promising results on benchmark datasets, evaluating on a code-mixed corpus with hyper parameter values found
to be the best on such benchmark datasets, as well as values in the same neighbourhood, results in unacceptable and impractical perplexity values for a NLM. 
We hope to explore the various strategies for developing a better language model of the datasets introduced in this work.







\bibliographystyle{IEEEtran}

\bibliography{references}

\end{document}